\documentclass[11pt]{article}
\usepackage{amsmath}
\usepackage{graphicx,psfrag,epsf}
\usepackage{enumerate}
\usepackage[numbers, sort&compress]{natbib}
\usepackage[hyphens]{url} % not crucial - just used below for the URL 

\usepackage{makecell}
\usepackage[table,dvipsnames]{xcolor}
\usepackage{amsmath}
\usepackage{amsthm}
\usepackage{amssymb}
\usepackage{bbold}
\usepackage{bm}
\usepackage{mathtools}
\usepackage{collcell}
\usepackage{dsfont}
\usepackage{rotating}
\usepackage{wrapfig}
\usepackage[utf8]{inputenc}
\usepackage{tikz}
\usepackage{lettrine}
\usepackage{dblfloatfix}
\usepackage[english]{babel} 
\usepackage{amsfonts}
\graphicspath{ {fig/} }
\usepackage[hyperfootnotes=false]{hyperref}

\usepackage{subfigure}

\usepackage{lipsum}
\usetikzlibrary{arrows,decorations.markings}
\usepackage{standalone}

\definecolor{gray}{rgb}{0.5,0.5,0.5}
\definecolor{red}{rgb}{0.8,0,0}
\definecolor{dred}{rgb}{0.5,0,0}
\definecolor{blue}{rgb}{0,0.1,1}
\definecolor{dblue}{rgb}{0,0.1,0.6}
\definecolor{cyan}{rgb}{0,0.5,.5}
\definecolor{dcyan}{rgb}{0,0.3,.3}

\definecolor{b}{rgb}{0,0,.8}	%%omega-blau
\definecolor{g}{rgb}{0,.6,0}	%%Tau-grün
\definecolor{n}{rgb}{0,0,0}	%%normal-schwarz
\definecolor{h}{rgb}{0.4,0.2,0.2}	%%hint
\definecolor{v}{rgb}{0.2,0.6,0}

\newcommand{\E}{{\mathbb E}}

\newcommand{\PP}{{\mathcal{P}}}

\newcommand{\bsX}{\boldsymbol X}

\newcommand{\bstheta}{\boldsymbol \theta}

\newcommand{\ov}\overline

\newcommand{\rig}\right
\newcommand{\lef}\left
\newcommand{\nf}\normalfont

\title{M5 Competition Uncertainty:
Overdispersion, \\ distributional forecasting, GAMLSS and beyond}
 \author{Florian Ziel\\ University of Duisburg-Essen}

\begin{document}

\maketitle

 \begin{abstract}
The M5 competition uncertainty track aims for probabilistic forecasting of sales of thousands of \emph{Walmart} retail goods. We show that the M5 competition data faces strong overdispersion and sporadic demand, especially zero demand. We discuss resulting modeling issues concerning adequate probabilistic forecasting of such count data processes. Unfortunately, the majority of popular prediction methods used in the M5 competition (e.g. \texttt{lightgbm} and \texttt{xgboost} GBMs) fails to address the data characteristics due to the considered objective functions. The distributional forecasting provides a suitable modeling approach for to the overcome those problems. The GAMLSS framework allows flexible probabilistic forecasting using low dimensional distributions. We illustrate, how the GAMLSS approach can be applied for the M5 competition data by modeling the location and scale parameter of various distributions, e.g. the negative binomial distribution. Finally, we discuss software packages for distributional modeling and their drawback, like the \texttt{R} package \texttt{gamlss} with its package extensions, and (deep) distributional forecasting libraries such as \texttt{TensorFlow Probability}.
 \end{abstract}

%  \begin{keyword}
% data cleaning \sep missing values \sep outliers \sep anomaly   
%  \end{keyword}

%keywords:
% M5 competition; probabilistic forecasting; GAMLSS; distribution modeling; overdispersion; count data
 
\section{Introduction}
% \citep{gneiting2011making}

The M5 competition aims on forecasting unit sales of thousands of \textit{Walmart} retail goods. 
This may be regarded predicting high-dimensional time series counting data that displays intermittency. For probabilistic forecasting in the uncertainty track it is important to address arising problems of dispersion and sporadic demand, especially for zero demand.
Overdispersion and sporadic demand are general phenomena observed for many count data time series, \cite{kolassa2016evaluating}. 
However, in retail forecasting, it can be partially explained. Zero demand may follow calendar pattern, especially
weekly, annual and holiday effects, but may be driven by promotion activities of the retailer as well. 
%sometimes .
Usually, the demand for a product directly transfers to sales.
However, sometimes there are zero sales even though there is demand for a certain product.
Those events might occur predominantly when a product is not-in-stock or displaced in the store.

% seasonality (weeky or yearly), or for other retailers promotional activity 

% Zero-inflated distributions are a suitable tool to address zero demand behavior that occurs for many products. 

In this manuscript, we discuss in detail the problem of intermittency in the context of the M5 competition data. 
We highlight that the adequate consideration of overdispersion substantially can substantially contribute to accurate forecasting models.
Additionally, addressing the sporadic zero demand
by utilizing zero-inflated distributions potentially improves forecasting results as well. 
Afterwards, we discuss aspects on distributional forecasting and
introduce the GAMLSS (generalized additive models location shape scale) framework
for distributional modeling
in the context of the M5 competition, \cite{stasinopoulos2007generalized, stasinopoulos2017flexible}.
We discuss how it can be used to efficiently incorporate all stylized facts in the data, 
like autoregressive, calendar effects, overdispersion, zero-demand, and external regressor effects.
A probabilistic forecasting application using GAMLSS for the M5 competition data using the \texttt{R} package \texttt{gamlss.lasso} is provided as well.
Finally, we discuss available 
software packages and their drawbacks for training distributional models, especially the \texttt{R} packages \texttt{gamlss}, \texttt{bamlss} and important extensions, and \texttt{TensorFlow probability} and related (deep) learning tools.

% 
% extension and the application of zero-inflated distributions in software packages. We also comment how the methodology can be used in popular statistical and machine learning packages for gradient boosting machines and artificial neural networks.

Please note that in the remaining part of the manuscript, we consider for illustration purpose the M5 competition data only on lowest hierarchical level for each item in each store and ignore further hierarchical aspects.

\section{Overdispersion in the M5 competition data}
 \begin{figure}
 \resizebox{.85\textwidth}{!}{
\includegraphics{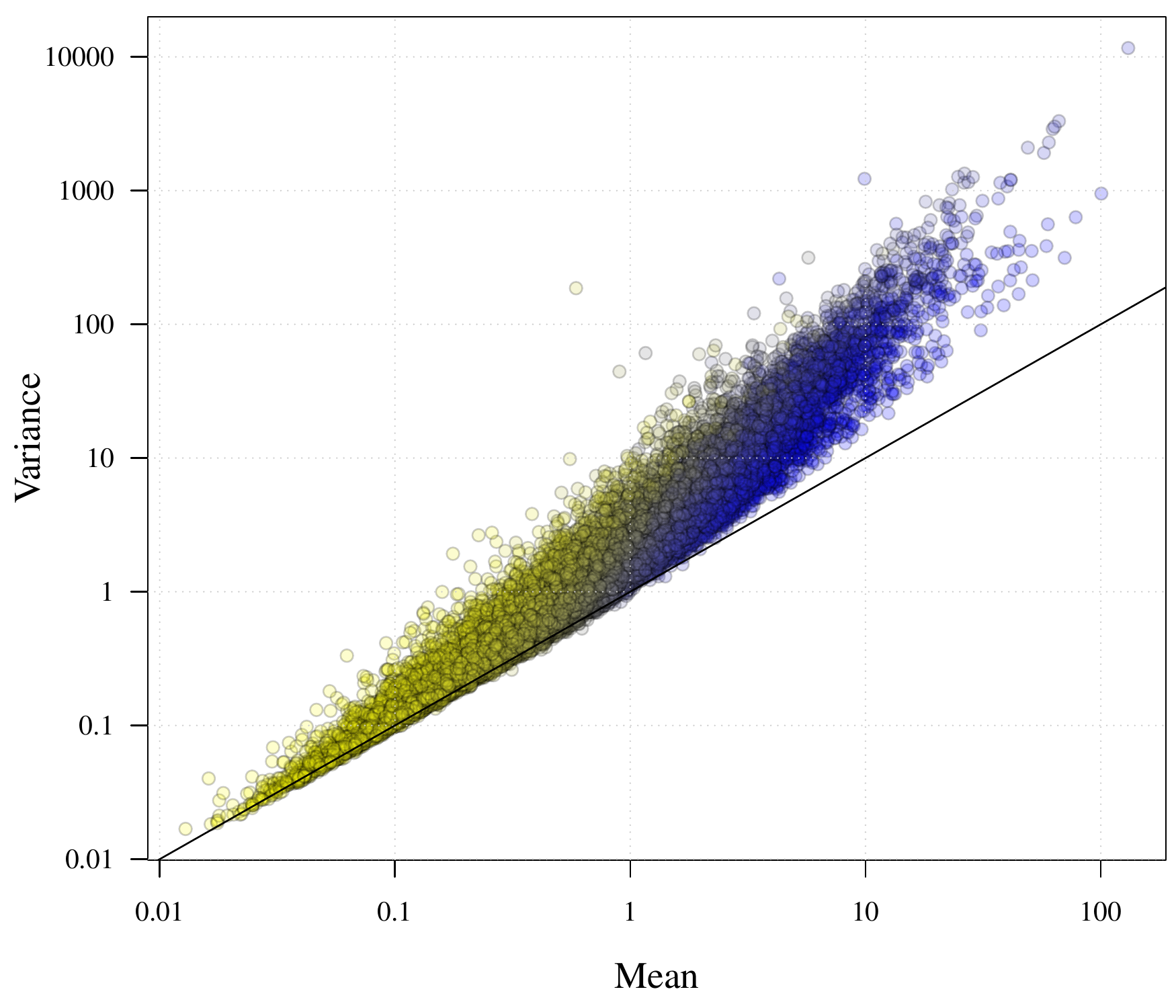}
}
\caption{Illustration of index of dispersion for 30490 items with black line indicating $\text{IOD}=1$ (perfect Poisson fit). The colors scheme indicated the proportion of zeros in the data. [yellow (only zeros) $\rightarrow$ blue (no zeros)]}
\label{fig_iod}
 \end{figure}

It is quite obvious to observe that the clear majority of the M5 time series are overdispersed. 
For counting data a typical measure for dispersion is the index of dispersion (IOD, also known as fano factor) 
which is the variance-mean ratio ($\text{IOD}=\sigma^2/\mu$) of the data.
 For the M5 data set the variance and the mean are visualized in Figure \ref{fig_iod} in a log scale. If the data would be perfectly Poisson distributed, then all points would lie close to the black line which corresponds to $\text{IOD}=1$. 
 We observe that in Figure \ref{fig_iod} almost all points lie clearly above the black line.  
%  But in Figure \ref{fig_iod} for almost all points we have a variance  the 
 Thus, the M5 competition data tends to be clearly overdispersed.

\begin{figure}
 \resizebox{\textwidth}{!}{
\includegraphics{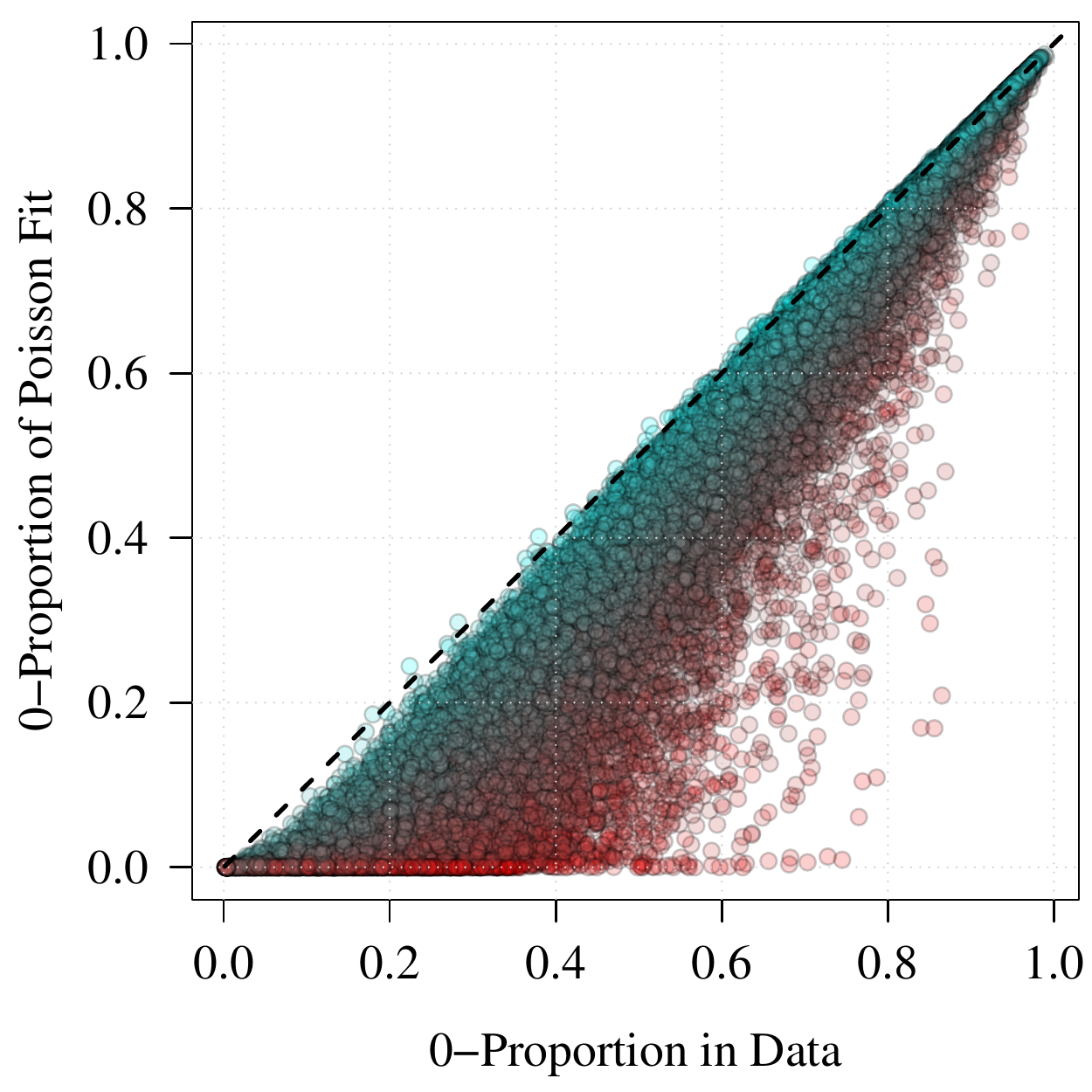}
\includegraphics{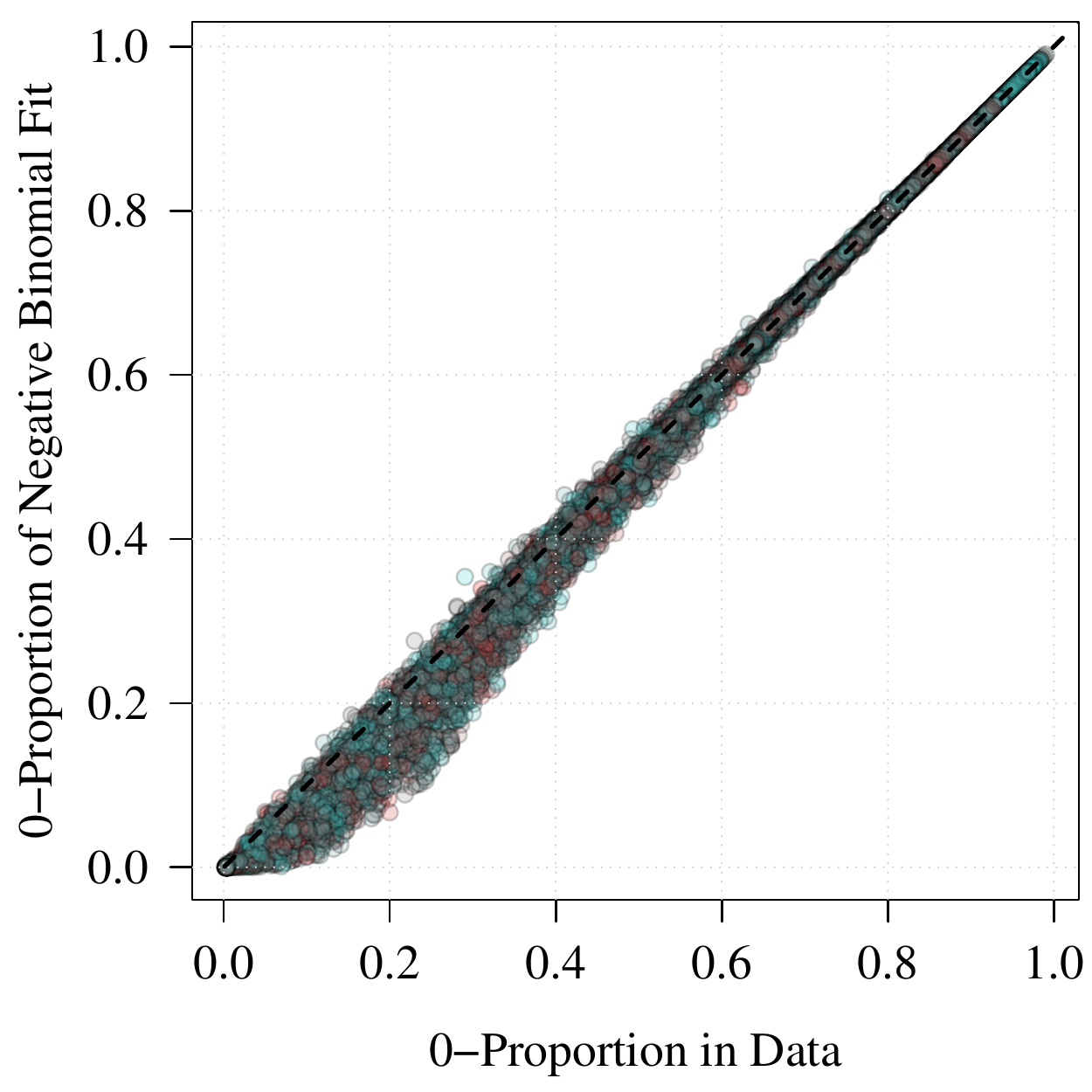}
}
\caption{Illustration of ratio of zeros in the data ($f(0)$) and corresponding estimate under a certain fit $\widehat{f}(0)$, Poisson (left) and negative binomial (right).
The colors scheme indicated the index of dispersion (IOD) in the data. [red (low) $\rightarrow$ cyan (high)]}
\label{fig_z0}
 \end{figure}

In statistics and probabilistic machine learning, one option to deal with overdispersion is to consider a distribution assumption that differs from the Poisson distribution. A common alternative is the negative binomial distribution. An alternative common feature which leads to overdispersion in the data could be zero inflation. 
To disentangle if both effects it is appropriate to have a 
look at the proportion of zeros in the data and compare it with an appropriate fit to the data. 
 Figure \ref{fig_z0}, shows corresponding maximum likelihood fits for a Poisson distribution and a negative binomial distribution.
 In the Poisson case we see the problematic fit again. The amount of zeros in the data is much higher than that modeled by the Poisson model. For the negative Binomial distribution the situation improved substantially, even though it is not perfectly captured as well. 
 Thus, it is a clear the consideration of overdispersion is crucially and can explain a big chunk of zero sales in the data. 
 
The clear presence of overdispersion might be a reason why some top scoring teams (the 1st place methodology of the accuracy track, 3rd place in the uncertainty track) of the M5 competition considered as an objective \emph{Tweedie} in \texttt{lightgbm}, a very popular gradient boosting machine (GBM) package, see \cite{makridakis2020m5}.
In the \texttt{lightgbm} implementation the Tweedie distribution is an overdispersed distribution with larger IOD compared to the Poisson distribution. The latter was used as objective by many other well scoring participants, and more the standard approach among the participants.
 
%Moreover, we see that the negative binomial model fit has a couple of results where the model predicts essentially 100\% zeros but .

 However in Figure \ref{fig_z0}, we still see  that the amount of zeros in the negative binomial fit is still a bit higher than in the data.
 This could be an indication for higher moment effects or zero-inflation in the data.
Plausible distribution functions that can address effects of higher moments include, for example, the beta-negative binomial distribution, but there are other options as well. 
To capture potential zero inflation, we need to add an appropriate parameter that explicitly models the zero component in the distribution. An example of such a distribution is the zero-inflated negative binomial distribution. 

In the M5 competition zero-inflated distributions were hardly discussed, even though there were several discussions in the forums about out-of-stock behavior and similar effects.
According to the \emph{kaggle} forum only one team reported that it used zero-inflated distributions in the competition for the final competition. This team scored 7th in the uncertainty track and considered a zero-inflated Poisson distribution.

% Whereas the problem of overdispersion is well understood by 

% not-in-stock behaviour for certain products
% .

% /home/florian/sciebo/PRIVATE/m5/proposal/z0pois.pdf
% /home/florian/sciebo/PRIVATE/m5/proposal/z0nb.pdf

% Simple 

%For the first time, it focused on series that display intermittency, i.e., sporadic demand including zeros.

% We analyse the role of intermittency and dispersion in the context of the M5 competition in detail. 

\section{Why distributional forecasting and GAMLSS?}

Of course, there are plenty of ways to generate probabilistic forecasts 
from a model as requested in the M5 competition, where prediction task focuses forecasting quantiles at the probabilities $\PP = \{0.005, 0.025, 0.165, 0.25, 0.5, 0.75, 0.835, 0.975, 0.995 \}$. 
However, it is still possible to disentangle all forecasting methods into two groups:
\begin{itemize}
 \item[i)] A forecasting method that predicts each $p$-quantile with $p\in \PP$ directly.
% uses a separate forecasting model 
 \item[ii)] A forecasting model that forecasts a distribution for the prediction target time series and evaluates the $p$-quantile for $p\in \PP$ of the predicted distribution.
\end{itemize}

% I
% The advantage of ii) in contrast to i) is that we only have to train one model. 
In i) quantile regression type objective functions are predominantly used in all forecasting methods independent of the model class (e.g. statistical time series, or machine learning type models like GBMs or neural networks).
If the approach i) or ii) is advantageous in the M5 competition setting is not clear. Both have advantages, 
e.g. i) models directly the object of interest, while ii) faces no problem concerning quantile crossing \cite{chernozhukov2010quantile}.
However, it is clear that the 
larger the set $\PP$ the more computational effort is required for approach i).
If we follow ii) then all quantiles are directly computed from the predicted distribution. The availability of the full predictive distributions is also preferable for decision making as it provides more information. For instance, we can compute convolutions of predicted densities which correspond to the density of the sum if we assume independence.

In general, there are 
two key methods to tackle ii), it could be done parametrically or 
non-parametrically. The latter approach was applied in demand forecasting competitions as well \cite{haben2016hybrid}, but tends to struggle for large data sets. Parametric approaches are usually easier to implement. 
As already mentioned, also in the M5 competition many participants applied a probabilistic forecasting methods using a parametric distribution objective. Predominantly, the Gaussian and Poisson distribution, but also the Tweedie distribution. This holds especially for those participants which applied gradient boosting machines with \texttt{lightgbm} or \texttt{XGboost}.

The problematic aspect of such an approach is that the vast majority of learning algorithms, especially in machine learning is only designed for one-parametric optimization problems. The single parameter is usually the location parameter of the underlying distribution. For instance,
 optimizing with respect to the $\ell_2$-loss is equivalent to maximizing likelihood estimation (MLE) for a normal distribution with parameterized mean parameter and constant variance. Similarly, minimizing the quantile loss for probability $p$ is equivalent to 
 MLE for a asymmetric Laplace distribution with parametrized location parameter, constant scale parameter and asymmetry parameter $p$. The $\ell_1$-loss is an important special case for $p=0.5$.
The Poisson loss maximizes with respect mean of the Poisson distribution assumption to a given certain offset. And also the more general Tweedie assumption only maximizes
with respect to the mean while the dispersion parameter of the Tweedie distribution has to be constant.
As pointed out in the introduction, it certainly helps to specify a dispersion parameter in the Tweedie assumption to improve to model fit and account for overdispersion compared to the Poisson approach. But this is likely not sufficient, because the assumption that all data points in the training of the forecasting model (potentially varying across time, item and store) can be adequately modeled using the same dispersion level is rather unrealistic.

To tackle the described problem one possible solution are probabilistic models that model not only the location parameter, but other aspects of the distribution. 
Distributional forecasting frameworks allow to model multiple parameters of a distribution, not only a single location parameter. 
The GAMLSS (Generalized Additive Models Location Shape Scale) approach  follows this path and considers the parametric option in method ii), see \cite{stasinopoulos2007generalized}. 
GAMLSS uses an optimization algorithm based on iterative weighted least squares (IWLS) to tackle the potentially very complex problem distributional forecasting problem. 
This can be a clear plus in contrast to direct likelihood optimization methods, mainly used in deep distributional forecasting \cite{salinas2020deepar, klein2020deep, chen2020probabilistic}. For more algorithmic details in Section \ref{sec_algorithm}. 

As pointed out in \cite{klein2015bayesian}, GAMLSS is nowadays used only as a word for IWLS based distributional forecasting (often in combination with the application of the \texttt{gamlss} package), sometimes it is also treated a synonym to distributional forecasting. This holds even if the considered parameters of the distribution parameters are usually not interpreted as location, scale and shape parameters. 
% For instance the when considering the zero-inflated Poisson distribution, the zero-inflation parameter is definitively not a scale parameter, but an inflation parameter. 
Still, one might argue that every distribution parameter that is not a location or scale parameter is a shape parameter. Anyway, independent of the wording it can be modeled using the GAMLSS approach.

%petropoulos2021forecasting
\section{Distributional forecasting meet the M5 competition}
% param
To introduce distributional forecasting for the M5 competition, we require some notations. 
Let $Y_{t,i,j}$ be the sales of item $i$ in store $j$ at time $t$ (in days as in the M5 competition) and $\bsX_{t}$ a vector of features that is available at time $t$.
$\bsX_{t}$ may contain external inputs and derived features, so e.g. lagged values of $Y_{t,i,j}$, item, store, calendar, Supplemental Nutrition Assistance Program (SNAP) and price information. In linear models we may add nonlinear transformations and interactions of mentioned inputs.

Additionally let $F_{\bstheta}$ be an $M$-parametric univariate distribution with parameter vector $\bstheta = (\theta_1, \ldots, \theta_M)$ which will be our distribution assumption for the data. This could be a continuous density like the normal, gamma or beta distribution, a counting density, like the Poisson or negative binomial distribution, or a mixture of both types, like the zero-inflated gamma distribution. Obviously, for the M5 competition counting densities are a natural choice, as we model counts.

Now, for a forecasting horizon $h$ we define the distributional model by
$Y_{t+h,i,j} \sim F_{\bstheta}$ with 
\begin{equation}
g_m(\theta_m) =  \sum_{n=1}^{N_m} f_{m,n}( \bsX_{t} ),
\label{eq_gamlss}
\end{equation}
invertible link functions $g_m$, and $N_m$ model component functions $f_{m,n}$ which depends potentially on all input features $\bsX_{t}$.
If $\theta_m$ takes values in $(-\infty, \infty)$, a typical assumption for $g_m$ is the identity. If $\theta_m$ is positive, then often $g_m = \log$ is considered, \cite{stasinopoulos2007generalized, stasinopoulos2017flexible, salinas2020deepar}.
The logarithm comes with nice property that is strictly monotonically increasing of on its support, but its inverse is the exponential function. This may lead to dangerous situations if $\log(\theta_m)$ is predicted with a large value the exponential function might led the forecast of the distribution parameter explode. 
However, there are alternatives to the $\log$.  %as proposed by
 For instance \cite{narajewski2020ensemble} propose the \text{logident} link function to handle those situations. It is defined by
\begin{equation}
 \text{logident}(x) = \log(x) \mathbb{1}(x\leq1) + (x-1)\mathbb{1}(x>1).
 %log(σ)1(σ≤1) + (σ−1)1(σ >1) .
 \end{equation}
This is the logarithm of $(0,1]$ with a linear function continuously glued on it at for $x\geq1$.

The flexibility of the distributional approach arises from the \emph{additive} components $f_{m,n}$. Obviously, if $N_m=1$ and  $f_{m,1}$ is linear, then we are in a standard linear model world.
GAMLSS was the first popular approach for efficient general distributional forecasting. 
Historically, GAMLSS was mainly developed for (smoothing) splines as
additive choice for $f_{m,n}$ to account for potential non-linearities in selected external inputs, \cite{rigby2005generalized}. However, there is no restriction on the model components. Nowadays, almost everything which popular in statistics and machine learning is utilized. For example, we have software solutions for different splines, GBMs and shallow and deep neural networks, esp. multi layer perceptrons.

The crucial aspect in any distributional forecasting model \eqref{eq_gamlss}
is its potential complexity. Highly complex models have 
high computational effort and problems with overfitting. 
Overfitting can be tackled basically in the way as with all standard approaches, e.g. by regularization (esp. lasso $\ell_1$ and ridge $\ell_2$ penalty), pruning, early stopping or drop-out depending on the models. 

However, in distributional forecasting settings overfitting requires more attention than in point forecasting frameworks. To get a better understanding why this is the case, consider the normal distribution (or if you wish the negative binomial distribution) which is parameterized by
$\mu$ and $\sigma^2$. The former is the expected value $\mu = \E[Z]$, the latter the variance $\sigma^2 = \E[(Z-\E[Z])^2] = \E[(Z-\mu)^2] $ of a random variable $Z$. It is easy to see that the variance $\sigma^2$ depends on $\mu$. So if we have a catastrophic fit in $\mu$ any model for $\sigma$ is obsolete. 
%, as it will definitively not model the expected squared distance to the expected value of $Z$. 
So a model for $\sigma$ faces the uncertainty from a model for $\mu$ on top of the standard uncertainties. Thus, the danger of overfitting is generally larger for $\sigma$ than for $\mu$. For higher moment effects this problem often gets even worse. As a results the complexity in well regularized distributional models usually decreases for higher moment effects.

% Obviously, the more complex the 
% structure of the model components $f_{n,m}$ the higher the computational demand.

\section{An illustrative GAMLSS example on M5 data}

Here, we want to illustrate how \texttt{gamlss.lasso} \cite{gamlsslasso} which is an add-on \texttt{R}-package to \texttt{gamlss}
can be used in the M5 competition setting. We apply it to all 3490 items for all 10 stores.
Yet, this is just an illustrative example, so the model is definitively not optimal, but it is reasonable. 

\begin{figure}
 \resizebox{\textwidth}{!}{
\includegraphics{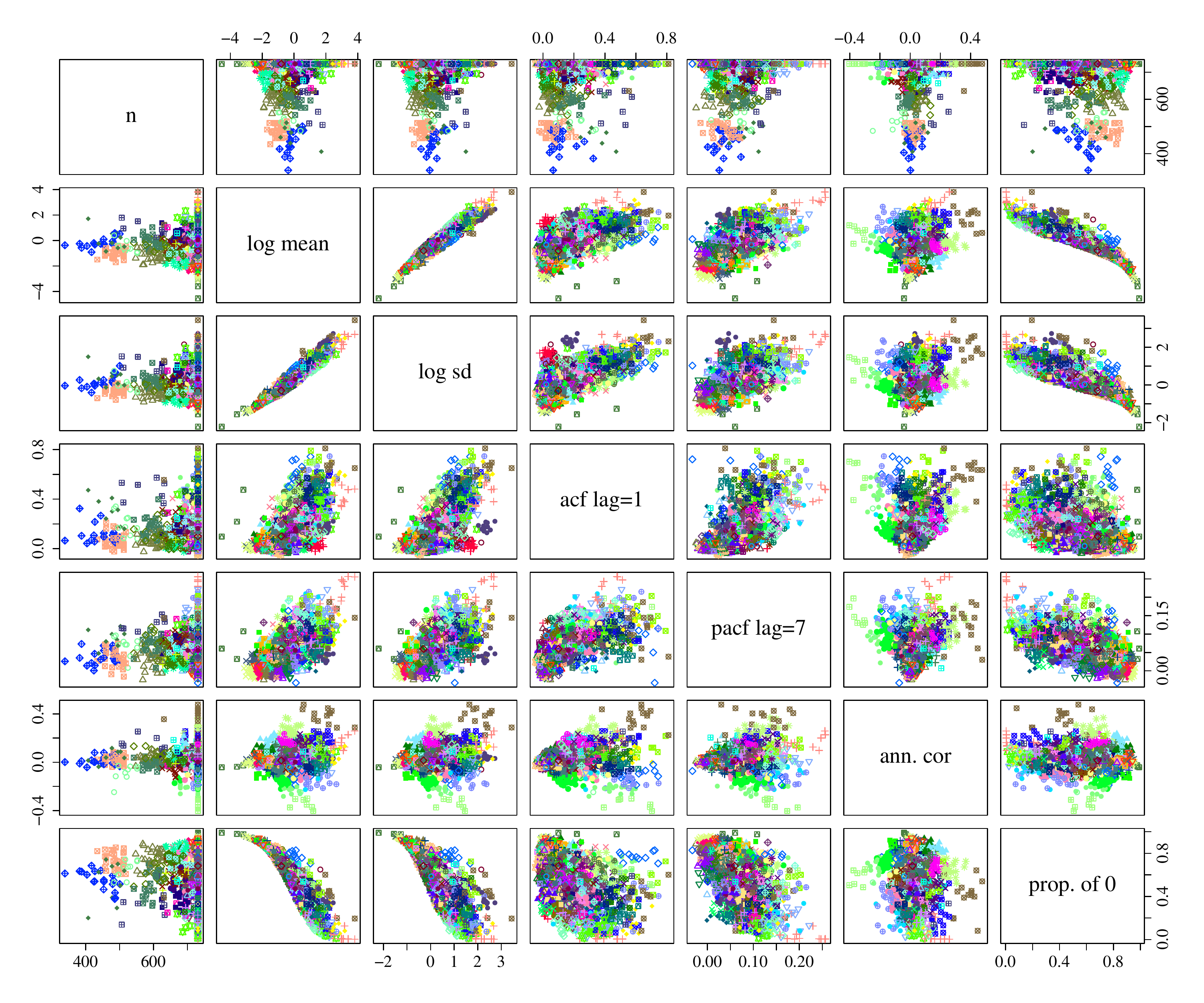}
}
\caption{All 3049 items, with 7 selected features average over all stores. 
The clusters are highlighted by different colors and symbols.}
\label{fig_cluster}
 \end{figure}

Due to the computational effort, we cannot apply the model to all 30490 products in a large probabilistic forecast model at the same time.  
Therefore, we consider only the last 2 years of data and apply a simple clustering method in advance to cluster the data into $100$ groups. We apply k-means on features for all items averaged across all stores. 
The considered features are: the sample size, the logarithm of the mean and standard deviation (see Fig. \ref{fig_iod}), the ACF at lag 1, the PACF at lag 7, the correlation of the time series with a 52$\times$7 days lagged rolling mean of 7 days, and the proportion of zeros in the data.
The clustering, results are presented in Figure \ref{fig_cluster}.
The clusters contain between 3 and 83 items with a median cluster size of 28.

Now, denote $Y_{t,\cdot,j}$ the average demand across all available items at time $t$ and store $j$, 
$Y_{t,i,\cdot}$ the average demand across all available stores at time $t$ and item $i$, and $Y_{t,\cdot,\cdot}$ the average demand across all available items and stores at time $t$. All those averages are taken within the considered cluster with $K$ items.
Then, we consider initially a linear model in each distribution parameter $\theta_m$ for $Y_{t+h,i,j}$ for $h=1$ given by
\begin{align}
g_m(\theta_m)  
= \beta_{m,0}&+\underbrace{ \sum_{k\in \{0,1, 6, 7\times52-1\}} \beta_{m,1,k} Y_{t-k,i,j}
+ \beta_{m,2,k} Y_{t-k,\cdot,j}
+ \beta_{m,3,k} Y_{t-k,i,\cdot}
+ \beta_{m,4,k} Y_{t-k,\cdot,\cdot}
}_{\text{lagged demand}} \nonumber \\
&+ \underbrace{ \sum_{k\in\{6,13,27,55\}}\beta_{m,5,k} \ov{Y}^{0:k}_{t,i,j}
+ \beta_{m,6,k} \ov{Y}^{0:k}_{t,\cdot,j}
+ \beta_{m,7,k} \ov{Y}^{0:k}_{t,i,\cdot}
+\beta_{m,8,k} \ov{Y}^{0:k}_{t,\cdot,\cdot}
}_{\text{rolling mean of lagged demand}} \nonumber \\
&+ \underbrace{\beta_{m,9} \ov{Y}^{0:6}_{t-7\times52,i,j}
+ \beta_{m,10} \ov{Y}^{0:6}_{t-7\times52,\cdot,j}
+\beta_{m,11} \ov{Y}^{0:6}_{t-7\times52,i,\cdot}
+\beta_{m,12} \ov{Y}^{0:6}_{t-7\times52,\cdot,\cdot}
}_{
\text{last years demand level}} \nonumber
\\
&+  \underbrace{\sum_{k=\in\{0,4,5,6\}} \beta_{m,13,k}\text{DoW}_t(k)}_{
\text{weekday effect}}
+  \underbrace{\sum_{k=0}^{K-1} \beta_{m,14,k}\text{item}_i(k)}_{
\text{item effect}}
+  \underbrace{\sum_{k=0}^9 \beta_{m,15,k}\text{store}_j(k)}_{
\text{store effect}}
\label{eq_gamlss_example}
\end{align}
with %lag set $\LL = $,
rolling mean $\ov{Y}^{k_1:k_2}_{t,i,j} = 
\frac{1}{k_2-k_1+1}\sum_{k=k_1}^{k_2} Y_{t-k,i,j}$,
day-of-the week dummy $\text{DoW}_t$ for Monday, Friday, Saturday and Sunday, such as item and store dummies. Note that the weekday, item and store information is considered categorically encoded. It is easy to observe that this model has $1+(4+4+1)\times 4+4+K+10=K+51$ model parameters, so depending on the cluster size around 100 model parameters for each distribution parameter. This is smaller than the total number of considered demand time series (number of items $\times$ number of stores). Therefore, the considered model \eqref{eq_gamlss_example}
 will be likely be still not perfect, especially interaction effect are likely missing.

%  Hence, we consider a model \eqref{eq_gamlss_example} including all quadratic interactions of equations \eqref{eq_gamlss_example}. This approach corresponds to standard tailor expansion of 2nd order. From a machine learning perspective, this is a feature space expansion by applying the kern trick with a quadratic kernel. The number of parameters increases significantly, to 
%  4451.\footnote{Some of the interaction effects are exactly zero and will be dropped, e.g. $\text{DoW}_t(0) \text{DoW}_t(1)$.}
%   This parameter space is relatively large, but relevant feature can be filtered efficiently by regularization, e.g. using lasso.
The recent article \cite{ULRICH2021831} applies GAMLSS on similar e-grocery data. It utilizes the counting densities of the Poisson and negative Binomial distribution, and the continuous Gaussian and gamma distributions.
We consider 7 counting distribution assumptions: Poisson, geometric, negative binomial, Waring, generalized Poisson, double Poisson and zero-inflated Poisson. This selection includes 1-parametric distributions (Poisson and Geometric) and the remaining distributions are 2-parametric. 
We do not consider  3- or 4-parametric distributions to limit the  computational effort. 
If we consider e.g. the negative binomial distribution then have $M=2$ and  
$\bstheta = (\theta_1, \theta_2)$ where $\theta_1$ is the location parameter and $\theta_2$ the scale parameter.
As both, $\theta_1$ and $\theta_2$ take values in $(0,\infty)$ we consider the (default) link functions $g_m=\log$. Note that all considered distributions are parameterized with parameters in $(0,\infty)$ and have the default $\log$ link function in \texttt{gamlss}, except the zero-inflation Poisson distribution. Here, the zero-inflation parameter has the logit link  functions, because it takes values in $(0,1)$.

We train the model using the last two years of data using \texttt{gamlss.lasso} in \texttt{R}. The lasso regularization helps to reduce the overfitting problem. The lasso tuning parameters are chosen based on minimal Hannan-Quinn information criterion (HQC) which 
is a compromise between the rather conservative BIC and the aggressive AIC. Next to the individual model, we also consider a model which selects always the distributional model with minimal HQC.
The training on the full data set (including tuning parameter selection) for the plain model \eqref{eq_gamlss_example} takes from a few seconds to a few minutes depending on the distribution and the cluster size. 

\begin{figure}
 \resizebox{\textwidth}{!}{
\includegraphics{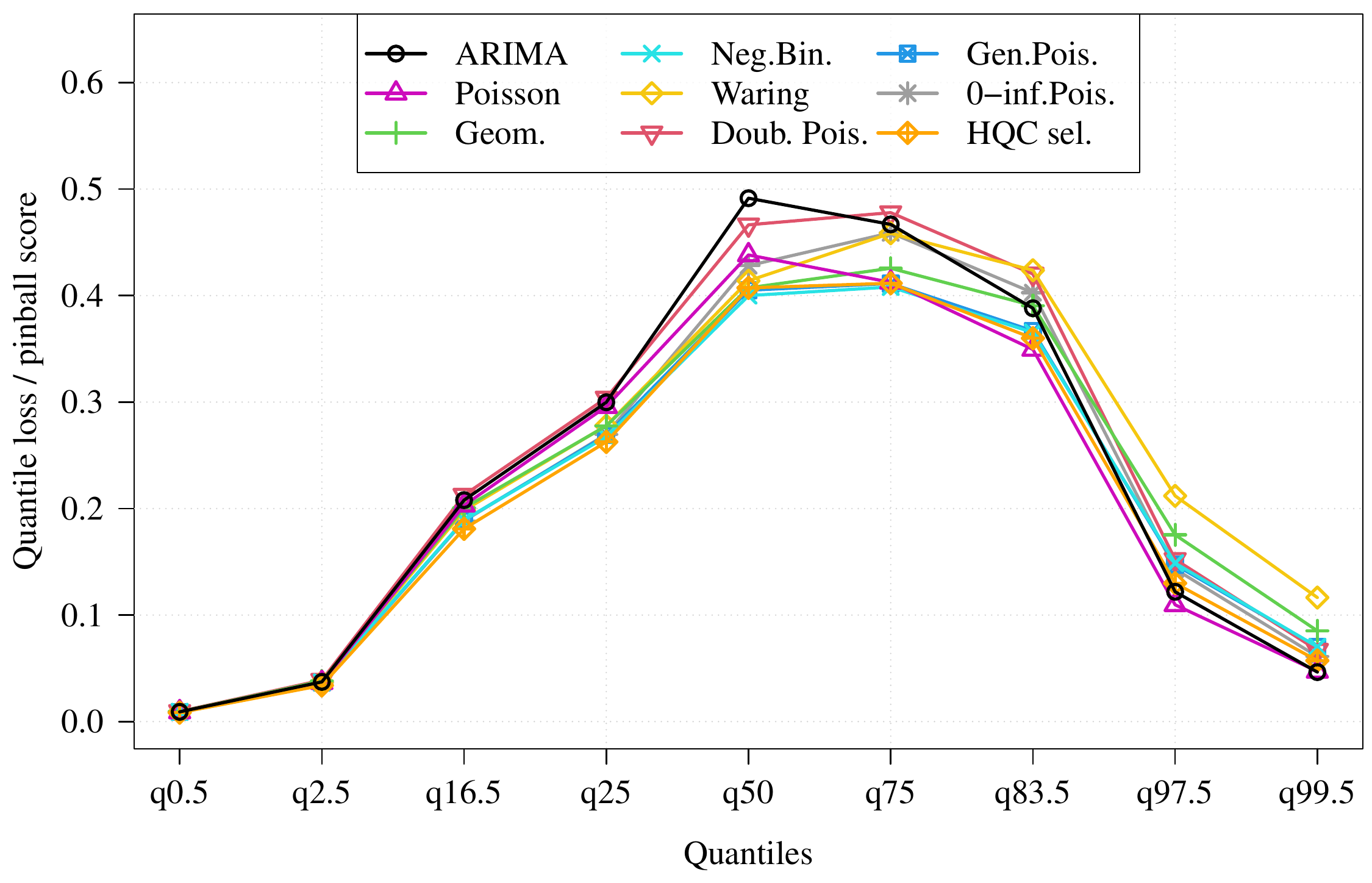}
}
\caption{Quantile/Pinball loss for the quantiles $\PP$ in the M5 competition for the considered distributions, as well as the ARIMA benchmark and the optimal HQC selection.}
\label{fig_pb}
 \end{figure}

% However,
% training the  model with interactions and 4451 parameters takes a couple of minutes on a standard computer using a single core.
The results are presented in Figure \ref{fig_pb} and Table 
\ref{tab_sum}. There we show, the results for all 7 distributional forecasting models, with the optimal HQC selected model and the ARIMA benchmark. The latter was the best performing benchmark in the M5 Uncertainty Competition. Table \ref{tab_sum} shows the average scores across the quantiles, and Figure \ref{fig_pb} the scores for all quantiles.
We observe that the Poisson, negative binomial and generalized Poisson distribution perform best, and around 8\% better than ARIMA. The major improvement comes from a better fit in the (upper) center quantiles at 50\% and 75\%. Table \ref{tab_sum} shows that the HQC selected solution yield more than 10\% improvement with respect to the ARIMA model. Thus, another 2\% gain in comparison to the negative Binomial model. Note that the best performing forecasts in the M5 competition had a corresponding improvement of around 12\% on the lowest hierarchical level. Still, compared to its structural simplicity 
of the linear model equation \eqref{eq_gamlss_example}
the performance  is quite remarkable,  even though price, SNAP, holidays, non-linear and interaction effects are not included.

Table \ref{tab_sum} also provides results concerning the HQC values of individual models. The negative binomial model which is the best individual model, has also best performance concerning the HQC, and has most often minimal HQC across all clusters. 
However, also the Waring, Generalized Poisson and zero-inflated Poisson distribution have relevant contributions to the HQC solution. This clearly indicates that there is not \emph{the} optimal distribution as the optimal distribution seems to depend on  data characteristics, a similar finding as in \cite{ULRICH2021}. For instance, for about 6\% of all items the zero-inflated Poisson distribution has minimal HQC. This indicated that only in about 6\% of the data has significant zero-inflation, and cannot be explained by overdispersion.

% for $h=1$.
% 
% 
% A plausible assumption 
% 
% 
% with link functions
% 
% 
% 
% The problematic part is that 
% 
% 
% However, as pointed out in Figure \ref{fig_z0}, we require more complex
% distributions to address both overdispersion potentially zero-inflation. 
% 
%  \begin{figure}
%  \resizebox{\textwidth}{!}{
% \includegraphics{out/gamlss02_item10001.pdf}
% \includegraphics{out/gamlss02_item10002.pdf}}
%  \resizebox{\textwidth}{!}{
% \includegraphics{out/gamlss02_item10011.pdf}
% \includegraphics{out/gamlss02_item10012.pdf}}
%  \resizebox{\textwidth}{!}{
% \includegraphics{out/gamlss02_item10021.pdf}
% \includegraphics{out/gamlss02_item10022.pdf}}
%  \resizebox{\textwidth}{!}{
% \includegraphics{out/gamlss02_item10031.pdf}
% \includegraphics{out/gamlss02_item10032.pdf}}
% \caption{M5 demand data of the first four items and two stores, with quantile of probability levels $\PP$ as used in the M5 competition for the last two years of data.}
% \label{fig_gamlss_example}
%  \end{figure}

%  Figure \eqref{fig_gamlss_example} shows fitted models for the first four items in the first two stores. 
% We illustrate the data together with fitted quantiles for probability levels $\PP$ used in the M5 competition.
% We observe that the model captures all relevant features.
% Also the calibration seems to be fine.

% latex table generated in R 4.1.0 by xtable 1.8-4 package
% Mon Jul  5 14:30:39 2021
\begin{table}[tbh!]
\centering
\setlength{\tabcolsep}{2pt}
 \resizebox{\textwidth}{!}{
\begin{tabular}{r|rrrrrrrrr}
  \hline
 & ARIMA & Poisson & Geom. & Neg.Bin. & Waring & Doub.Pois. & Gen.Pois. & 0-inf.Pois. & HQC-sel. \\ 
  \hline
av. QL/PB & \cellcolor[rgb]{1,0.67,0.5} $0.2299$ & \cellcolor[rgb]{0.827,1,0.5} $0.2112$ & \cellcolor[rgb]{1,0.87,0.5} $0.2233$ & \cellcolor[rgb]{0.785,1,0.5} $0.2105$ & \cellcolor[rgb]{1,0.5,0.547} $0.2386$ & \cellcolor[rgb]{1,0.5,0.55} $0.2388$ & \cellcolor[rgb]{0.852,1,0.5} $0.2116$ & \cellcolor[rgb]{1,0.905,0.5} $0.222$ & \cellcolor[rgb]{0.5,0.9,0.5} $0.2058$ \\ 
  Improv. [\%]& \cellcolor[rgb]{1,0.67,0.5} $0$ & \cellcolor[rgb]{0.83,1,0.5} $8.11$ & \cellcolor[rgb]{1,0.87,0.5} $2.87$ & \cellcolor[rgb]{0.789,1,0.5} $8.41$ & \cellcolor[rgb]{1,0.5,0.547} $-3.78$ & \cellcolor[rgb]{1,0.5,0.55} $-3.87$ & \cellcolor[rgb]{0.854,1,0.5} $7.94$ & \cellcolor[rgb]{1,0.905,0.5} $3.44$ & \cellcolor[rgb]{0.5,0.9,0.5} $10.48$ \\ 
  HQC & - & \cellcolor[rgb]{1,0.503,0.5} $2.2672$ & \cellcolor[rgb]{1,0.985,0.5} $2.1626$ & \cellcolor[rgb]{0.5,0.9,0.5} $2.1019$ & \cellcolor[rgb]{1,0.5,0.55} $2.2862$ & \cellcolor[rgb]{1,0.547,0.5} $2.2592$ & \cellcolor[rgb]{0.833,1,0.5} $2.1326$ & \cellcolor[rgb]{1,0.55,0.5} $2.2585$ & - \\ 
  best HQC [\%]& - & \cellcolor[rgb]{1,0.5,0.536} $1$ & \cellcolor[rgb]{1,0.5,0.55} $0$ & \cellcolor[rgb]{0.5,0.9,0.5} $35$ & \cellcolor[rgb]{0.671,0.986,0.5} $32$ & \cellcolor[rgb]{1,0.5,0.55} $0$ & \cellcolor[rgb]{1,0.964,0.5} $22$ & \cellcolor[rgb]{1,0.686,0.5} $10$ & - \\ 
  best wHQC [\%] & - & \cellcolor[rgb]{1,0.5,0.549} $0.1$ & \cellcolor[rgb]{1,0.5,0.55} $0$ & \cellcolor[rgb]{0.5,0.9,0.5} $35.5$ & \cellcolor[rgb]{0.832,1,0.5} $29.6$ & \cellcolor[rgb]{1,0.5,0.55} $0$ & \cellcolor[rgb]{0.855,1,0.5} $29.2$ & \cellcolor[rgb]{1,0.561,0.5} $5.7$ & - \\ 
  \hline
  \end{tabular}

}
\caption{Summary results for all considered model, in the M5 competition for the considered distributions, as well as the ARIMA benchmark and the optimal HQC selection. The table shows, the average quantile loss (=pinball score), the corresponding improvement to ARIMA, the HQC value, percentage of having minimal HQC for all clusters, and cluster size weighted percentage of having minimal HQC for all clusters.}
\label{tab_sum}
\end{table}

\section{Algorithmic issues for distributional forecasting} \label{sec_algorithm}

In this section, we discuss algorithmic issues on distributional forecasting, especially GAMLSS in more detail.
As already mentioned distributional models \eqref{eq_gamlss} can be very complex an lead to high computational costs. Therefore, computational aspects are highly relevant.

The GAMLSS procedure has a specific elements in the optimization that are usually beneficial to reduce computational costs. 
This is the iterative weighted least squares (IWLS) based backfitting procedure, \cite{rigby2005generalized, klein2015bayesian}.
Here, the Rigby and Stasinopoulos (RS), the 
Cole and Green (CG) algorithm such as Bayesian inference using
Markov chain Monte Carlo (MCMC) simulation techniques 
are available in practice. 

The RS procedure cycles iterative across the $M$ distribution parameters, by fixing the other ones. So first we optimize $\theta_1$ (to some extent), then $\theta_2$, $\ldots$ until we reach $\theta_M$ and then start again at $\theta_1$ until convergence or any other stopping criteria is reached. 
This can be faster than a global optimization which models all model parameter at once. This holds particularly, if the distribution parameters are orthogonal \cite{stasinopoulos2007generalized}, which is e.g. the case for the negative binomial distribution, if correctly parameterized.
Moreover, the danger of getting stuck in a bad local optimum is lower compared to a global approach if the distribution parameters are ordered suitably  (usually location, scale, shape is suitable).
Note that the cyclic procedure can be applied for methods that do not rely on IWLS but perform direct likelihood optimization as well. The consequences are the same. 

The IWLS approach for solving distributional regression problems comes with
all pros and cons for least square methods.
One drawback is that this works only for random variables with finite variance. This, for very heavy tailed data IWLS is not available.
Here, all methods that optimize the likelihood directly are preferable.
On the other hand, it allows us to utilize specialized least squares optimization algorithms for the model components in \eqref{eq_gamlss}.

\section{Software for distributional forecasting
and further developments}

Finally, we discuss available software packages 
for solving probabilistic forecasting problems as they occurred in the M5 competition. 
Further, we discuss briefly open research and implementation questions regarding distributional forecasting and GAMLSS for large data sets.

As already mentioned, the \texttt{gamlss} package is available in  \texttt{R}, \cite{stasinopoulos2017flexible}. It comes along with several additional packages, most notable 
 \texttt{gamlss.dist} with plenty of additional distributions, including zero-inflated distribution,
\texttt{gamlss.add} with additional smoothing components like 
artificial neural networks and decision trees, and
\textbf{gamlss.lasso} for efficient high-dimensional lasso type regression models. %,
There are also related \texttt{R} packages, like the \texttt{gamboostLSS} which focuses on gradient boosting machine applications, which is suitable for relatively large data sets. The package \texttt{bamlss} estimates GAMLSS models using Bayesian methods and has full distribution support from \texttt{gamlss.dist}.
% Also the efficient \texttt{gam} package provides some restricted support for GAMLSS type frameworks. 
% However, there is much less variety concerning smooting components and distributions available.

GAMLSS software was originally developed in \texttt{R} and designed only for small and medium sized date sets with only a couple hundreds or thousands of observations. Many, features which are useful for data analytics of (extremely) large data sets, like sparse matrix support, are hardly available for the \texttt{R} packages on GAMLSS. Only the add on \texttt{gamlss.lasso} package to \texttt{gamlss} which is designed for high-dimensional problems has some sparse matrix support, as well as \texttt{bamlss}. 
Here, further development is required, for instance in the direction of sparse matrix support for handling large sparse data sets.

Another important stream of available software for distributional modeling and forecasting is \texttt{TensorFlow probability} (\texttt{TFP})
which is available for \texttt{python}, but can be used in \texttt{R} as well, \cite{dillon2017tensorflow}.
\texttt{TFP} supports so called distribution layers 
which corresponds to the GAMLSS distribution modeling, \cite{klein2015bayesian}.
As \texttt{TFP} comes with huge support for neural network models, e.g. using \texttt{keras}, it allows for extremely complex probabilistic forecasting models using deep learning. The amount of available distributions is still limited and smaller than in \texttt{gamlss.dist}. However, this might change in the next years due to the increasing popularity of \texttt{TFP}, as pointed out by \cite{yang2020deep, rugamer2021deepregression} who considered deep neural networks for distributional regression problems.
Next to \texttt{TFP}, there is more deep learning software that supports distributional forecasting (see \cite{petropoulos2021forecasting} Sec. 2.7.9. for a broader discussion). Popular examples are the DeepAR and N-BEATS model implemented in the \texttt{GluonTS} library, see \cite{salinas2020deepar, oreshkin2019n}. However, at the moment only a few probabilistic distributions are supported. 
% However, due to the diversity of well established models, it is slightly easier to use than \texttt{TFP}. Still, in both cases hyperparameter tuning
% is crucial.
% Another key difference in the solution of the underlying distribution problem. \texttt{TFP} solves a global optimization problem for all $M$ parameter equations jointly, thus aiming for finding the global optimum. In contrast, the original GAMLSS approach (as considered in \texttt{gamlss}) cycles iteratively through the $M$ parameter equations of \eqref{eq_gamlss}. This tends to lead to more robust and fast model training. So it would be beneficial for the \texttt{TFP} community to go towards such training methods when distribution layers are involved.

Moreover, to our knowledge, all software options primarily designed for deep learning like \texttt{TFP} have no training framework available that is based on the IWLS backfitting algorithm - even though batchwise IWLS algorithms can be implemented. In general, so far only \texttt{bamlss} supports both optimization frameworks, the IWLS (the default method) and direct likelihood based optimization using boosting techniques.

As dicussed both optimization frameworks (IWLS based and direct likelihood optimization) have their pros and cons resulting in different forecasting performance. It may have a big impact on the computation time as well. To illustrate this in an example, we consider a linear model in each distribution parameter $\theta_m$ where we want to apply lasso regularization. As mentioned the IWLS allows the efficient integration of least squares based algorithms. For instance the \texttt{gamlss.lasso} uses for the $\ell_1$ regularization the very fast coordinate descent algorithm implemented in \texttt{glmnet} \cite{friedman2010regularization}. In contrast, \texttt{bamlss} uses no specialized optimization, leading to huge computational differences even though both utilize the IWLS approach. 
To get an intuition for the magnitude of differences, we conducted a simulation study that is provided in the online Appendix.
We consider a normal distribution with 50 potential parameters in $\mu$ and $\sigma$ and 1000 observations. We estimate/train the lasso models on the same exponential tuning parameter grid of size 100 and perform BIC tuning.
The \texttt{gamlss.lasso} takes approximately 0.3 seconds, \texttt{bamlss} and \texttt{TFP} require around 20 and 70 seconds. 
%This huge difference is mainly caused by the integration of specialized.

% Deepregression \cite{rugamer2021deepregression}

Finally, we want to mention that there was the 
\texttt{XGBoostLSS} project started by \cite{marz2019xgboostlss}
which aimed for bringing popular GBM methods, such as \texttt{XGBoost} and \texttt{lightgbm}, in the GAMLSS world.  But is seems to be not continued since 2019. However, the popularity of the GBMs in the M5 competition indicated a high demand for efficient distributional GBM methods, beyond the \texttt{gamboostLSS} \texttt{R}-package.

\bibliographystyle{apalike} 
\bibliography{ref}

\end{document}